\documentclass{sig-alternate-10pt}
\usepackage{graphicx}
\usepackage{epstopdf}
\usepackage{float}
\usepackage{balance}
\usepackage{comment}
\usepackage{amssymb, amsmath}
\usepackage{multirow}
\usepackage{xspace}
\usepackage{xcolor, color, soul}
\usepackage{url}
\usepackage{amsmath}
\usepackage[export]{adjustbox} %niru

\newcommand{\new}{\vskip 0.75em plus 0.1em minus .4em}

\newcommand{\squishlist}{
 \begin{list}{$\bullet$}
  {  \setlength{\leftmargin}{1.3em}
    } }

 \newcommand{\squishend}{
  \end{list} }

\usepackage{soul,color}

\graphicspath{{./}}

\usepackage{titlesec}
\titlespacing{\section}{0pt}{*4}{*1.5}

\title{IMUOptimize: A Data-Driven Approach to Optimal IMU Placement for Human Pose Estimation with Transformer Architecture}

\numberofauthors{1} 
\author{
\alignauthor
Varun Ramani, Hossein Khayami, Yang Bai, Nakul Garg, Nirupam Roy \\
\vspace{0.2in}
      \affaddr{University of Maryland, College Park}
      \vspace{0.5in}
}

\begin{document}
\begin{sloppypar}

\maketitle
\begin{abstract}
\new
This paper presents a novel approach for predicting human poses using IMU data, diverging from previous studies such as DIP-IMU, IMUPoser, and TransPose, which use up to 6 IMUs in conjunction with bidirectional RNNs. We introduce two main innovations: a data-driven strategy for optimal IMU placement and a transformer-based model architecture for time series analysis. Our findings indicate that our approach not only outperforms traditional 6 IMU-based biRNN models but also that the transformer architecture significantly enhances pose reconstruction from data obtained from 24 IMU locations, with equivalent performance to biRNNs when using only 6 IMUs. The enhanced accuracy provided by our optimally chosen locations, when coupled with the parallelizability and performance of transformers, provides significant improvements to the field of IMU-based pose estimation.

\end{abstract}

\section{Introduction}

Human pose estimation is essential for applications in animation, gaming, healthcare, and autonomous driving. Recent integration of full-body motion capture technologies like Xsens \cite{xsens} has expanded possibilities in gaming, fitness, and rehabilitation. Despite benefits, consumer adoption remains limited due to inconveniences associated with retrofitting homes or wearing specialized suits.

Traditionally dominated by vision-based techniques utilizing cameras, pose estimation faces challenges in environments with occlusions, lighting variations, or incomplete subject visibility. Recent computer vision advancements have improved accuracy, but challenges persist in occlusion and constrained environments.

An alternative involves leveraging Inertial Measurement Units (IMUs) in everyday devices, including smartphones, smartwatches, and wearables like activPAL\cite{paltechnologiesltd}. This project explores infrastructure-free body pose estimation using IMU data. IMUs, measuring rotational velocity and linear acceleration, offer advantages over vision-based methods, being less affected by environmental factors and providing real-time pose information. Highly portable and suitable for various applications, IMUs are promising for fitness tracking, rehabilitation, and virtual reality.

The motivation for using IMUs in pose estimation stems from their ability to capture motion data directly from the subject, thus providing a high degree of freedom and flexibility. This is particularly beneficial in scenarios where camera-based systems are impractical or intrusive. Furthermore, the proliferation of consumer devices equipped with IMUs, such as smartphones and smartwatches, opens up new possibilities for accessible and ubiquitous pose estimation solutions.

In this project, we push the boundaries of pose estimation by delving into the realm of sparse Inertial Measurement Unit (IMU) configurations \cite{yi_transpose_2021, yi_physical_2022}, endeavoring to reconstruct precise user poses using a significantly reduced number of sensors compared to conventional approaches \cite{xsens}. This endeavor introduces a unique set of difficulties stemming from the inherent ambiguity of sparse IMU data, where a given set of IMU readings may correspond to a myriad of potential poses. The previous works on using sparse IMUs such as DIP-IMU\cite{huang_deep_2018}, TransPose \cite{yi_transpose_2021}, and PIP \cite{yi_physical_2022} intuitively selected 6 joints to place sensors. However, our methodology goes beyond traditional strategic  methods, incorporating an LSTM to extract more nuanced information from the sparse IMU data. Using model interpretation tools such as \cite{captum}, we find the joints that most contribute to the pose estimation and cherry-pick them for our model. Finally, we develop a novel network based on the transformer and investigate its potential as a replacement for the LSTM.

\section{Related works}

We begin by examining the Deep Inertial Poser (DIP-IMU) presented by Huang et al. \cite{huang_deep_2018}. This groundbreaking work introduces a method for real-time 3D human pose reconstruction using a minimal set of six body-worn Inertial Measurement Units (IMUs). The core innovation of DIP-IMU lies in its effective use of deep learning, particularly a bidirectional Recurrent Neural Network (RNN), to overcome the challenge of inferring complex human poses from sparse IMU data. By leveraging synthetic data generated from extensive motion capture databases, DIP-IMU trains its model to accurately predict human poses in a variety of real-world scenarios. This approach not only advances the field of pose estimation by reducing the need for extensive sensor setups but also demonstrates significant improvements in terms of accuracy and computational efficiency over previous methods. 

Building on the work done in DIP-IMU, TransPose \cite{yang2021transpose} attempts to predict translations as well as poses. This innovative approach represents a significant advancement in the realm of IMU-based pose estimation, as it not only captures the static poses but also the dynamic movements and global translations of the human body. The methodology employed by TransPose is noteworthy for its multi-stage network architecture, which systematically breaks down the task of pose estimation into more manageable sub-tasks. This design enhances the system's ability to interpret sparse IMU data, leading to more accurate and granular pose reconstructions. TransPose's ability to estimate global translations sets it apart from its predecessors. It utilizes a fusion of two complementary methods: a foot-ground contact estimation based on the IMU measurements, and a root velocity regressor that predicts the local velocities of the root in its coordinate frame. The combination of these methods allows for a more robust and precise estimation of the body's movement in space, which is a critical aspect often overlooked in traditional pose estimation systems.

Finally, IMUPoser \cite{mollyn_imuposer_2023} builds on the foundations laid by DIP-IMU and TransPose, pioneering an approach that integrates inertial data from widely-used consumer devices such as smartphones, smartwatches, and earbuds. IMUPoser's algorithm is adaptable, capable of effectively utilizing various configurations of IMUs, regardless of their placement on the body. By prioritizing user convenience and the practicality of device availability, IMUPoser addresses key challenges in the field of motion capture. 

However, each of these papers presents the same opportunity for advancement - IMU locations are either chosen arbitrarily, in the case of DIP-IMU and TransPose, or somewhat inherited from the nature of the system, as in the case of IMUPoser. We will build on the pioneering work of these papers to try and figure out exactly where we can place our IMUs to achieve higher accuracy with the same, minimal number of IMUs.

\section{Preliminaries}

First, we should define \textbf{pose estimation}. Pose estimation is the task of mapping some data $X$, in this case IMU data, to a set of $n$ (usually 24) joint rotations $\theta^*$. Take a moment to think about why all possible poses can be uniquely represented by a set of joint rotations (and why information like joint positions is redundant).

To achieve this task, Inertial Measurement Units are central to our study. These devices fuse an accelerometer and gyroscope to measure linear and angular motion. A sample from an IMU contains 3D acceleration and rotation - so acceleration in $x$, $y$, and $z$ directions, along with $roll$, $pitch$, and $yaw$ rotations. For reasons discussed later, we'll represent the gyroscope's output as a $3x3$ \textit{rotation matrix}, so it will require 9 scalars to represent as opposed to just 3. So in our work, an IMU sample is a vector $\in \mathcal{R}^{12}$.

Actual poses will be represented and visualized computationally using the SMPL model, which is used to represent the human body in 3D. This parametric model, built from a comprehensive database of body scans, enables the creation of versatile and realistic human figures. 

We'll train our neural network on the AMASS dataset \cite{AMASS:2019}, a rich collection of motion capture data. It provides sequences of poses that we can use as labels to our model.

The heart of our project lies in processing and interpreting the orientation and acceleration data from IMUs. This time-series data, characterized by its sequential nature and temporal dependencies, is the bedrock upon which our predictive models are built. Our objective is to predict rotation data, but instead of using Euler angles, which are prone to gimbal lock and singularities, we opt for 3x3 rotation matrices. These matrices provide a more stable and comprehensive representation of three-dimensional rotations, crucial for accurate pose estimation.

\new

\section{System design}

\new

To recap, we're attempting to apply a data-driven approach to figure out optimal IMU positions. More formally, our objective is to devise a function \( F(X) \rightarrow 
\theta^* \) that maps IMU data to a set of joint rotations, representing the 
actual pose. Here, \( X \) represents a subset of the available 24 joints on the SMPL skeleton, 
specifically those most pertinent to our task.

The primary question is: How do we determine the most relevant IMUs from the 24 joints available? This query may initially seem akin to a dimensionality reduction 
problem. However, unlike Principal Component Analysis (PCA) that transforms 
high-dimensional data into a lower-dimensional space, our goal is to ascertain 
which dimensions (IMUs) can be discarded or not collected altogether.

Given this distinction, we require an alternative approach to discern which 
among the 24 possible SMPL joints best explain variance in the pose labels. Common 
methods like p-value testing, often used in libraries like \texttt{sklearn}, are 
impractical for our needs. P-values, indicating the likelihood of 
feature-induced variance in output, are more suited to simpler models and not to 
the large-scale neural networks we intend to employ.

Therefore, our methodology involves:
\begin{enumerate}
    \item Predicting poses using all 24 joints.
    \item Quantitatively assessing which of these IMUs contribute most 
    significantly to this task.
    \item Excluding non-contributory IMUs.
\end{enumerate}

\subsection{Data}

In our study, we focus on mapping Inertial Measurement Unit (IMU) data, specifically accelerations and rotations, to full body poses. This approach is crucial in understanding and interpreting the movement dynamics captured by IMU sensors.

Interestingly, the converse of the process is quite easy to calculate. Using a formula derived from TransPose \cite{yang2021transpose}, we can calculate the acceleration data between two poses - i.e. we can easily find IMU data given poses. This flexibility allows for the utilization of completely synthetic data in our analysis. The formula is given by:
\begin{equation}
    a[i] = (v[i] + v[i + 2] - 2 \cdot v[i + 1]) \cdot 3600
\end{equation}
where \( v \) represents an array of frames, and \( a[i] \) is the acceleration at frame \( i \).

Naturally, the rotation of each joint is equivalent to the rotation of the corresponding synthetic IMU.

The base dataset for our study is AMASS \cite{AMASS:2019}, an aggregation of numerous motion capture datasets. AMASS provides data in the SMPL format, a mesh model that demonstrates how a virtual "body" is contorted by a given set of joint rotations. The SMPL model is particularly significant as it contains a \( 24 \times 6890 \) matrix that transforms 24 joints into the 6980 vertices of the model. By running an argmax operation on each of the 24 weight sets, we can identify the vertex that most closely corresponds to each joint.

 \begin{table}[h]
    \centering
    \begin{tabular}{|c|l|c|l|}

\hline
Joint Index & Joint Name & Joint Index & Joint Name \\
    \hline
0           & Pelvis     & 12          & Neck       \\
1           & L Hip      & 13          & L Collar   \\
2           & R Hip      & 14          & R Collar   \\
3           & Spine1     & 15          & Head       \\
4           & L Knee     & 16          & L Shoulder \\
5           & R Knee     & 17          & R Shoulder \\
6           & Spine2     & 18          & L Elbow    \\
7           & L Ankle    & 19          & R Elbow    \\
8           & R Ankle    & 20          & L Wrist    \\
9           & Spine3     & 21          & R Wrist    \\
10          & L Foot     & 22          & L Hand     \\
11          & R Foot     & 23          & R Hand    \\
    \hline
    \end{tabular}
    \caption{Mapping of SMPL Joint Indices to Joint Names}
    \label{tab:joint_mapping}
\end{table}

Once we have vertices, which are points on the SMPL mesh, and their corresponding joints, we can compute IMU data on a per-joint basis by applying the above formula to the motion of the vertices between frames.

\subsection{The LSTM Model}

In addressing this problem, it is crucial to understand that we are dealing with a time series issue where isolated pose frames are insufficient. To synthesize IMU data effectively, one must consider the sequence of frames and their context within the surrounding frames. This calls for a model that ingests a series of \(X\) frames of IMU data and outputs \(X\) corresponding poses.

Long Short-Term Memory (LSTM) networks are exceptionally suited for this challenge due to their ability to process sequences, retain information about previous frames, and apply this context to new ones. A prediction at any frame is a weighted consideration $f(h, x)$ that considers both the current frame $x$, along with the model's ``memory'', the hidden state $h$.

Below, we have provided our model architecture as a list. Here, $N$ is the number of sensors the model uses - it will either be 24 or 6.
\begin{enumerate}
    \item Linear Layer: Transform from shape $N * (3*3 + 3)$ to $1024$.
    \item LSTM Layer: Accept shape $1024$ then feed it through 2 recurrent layers, each with hidden size $1024$. Produce output of shape $2 * 1024$.
    \item Linear Layer: Transform from shape $2 * 1024$ to final output shape $24 * (3*3)$.
\end{enumerate}

RNN training was executed over 5 epochs, where each epoch took approximately 20 minutes to run on an RTX3090. Counterintuitively, note that we actually \textit{want} to overfit the model - we do not implement any sort of regularization, since our desired result is actually the model conditioning heavily on a small set of features, as opposed to using all the features equally. 

\subsection{Model Interpretation}
To figure out which features are most important, we need to leverage a model interpretation technique. Simpler models, like random forests and gradient-boosted trees, \cite{breiman_random_2001} are easier to interpret, but since we work with deep learning models, it can be a little harder to understand what goes into the network's prediction.

We apply Captum, a powerful library for model interpretability in PyTorch. Captum offers various methods for attributing model outputs to input features, aiding in the understanding of the model's decision-making process. Given that our model input is a tensor of size 288, representing 24 joints with 3-axis acceleration and 3x3 elements for rotations, we employ Captum's feature masking capabilities to analyze the importance of individual IMUs. 

Feature masking involves selectively hiding or masking certain features during model evaluation to observe their impact on predictions. Notably, since only a subset of methods in Captum supports feature masks, we focus on those that apply to our scenario. We tried Shapley value, Shapley value sampling, and feature ablation - we found that feature ablation was the most convenient to use and provided the best results.

Feature ablation is known as a \textbf{perturbation approach}: it systematically replaces given features or groups of features with a baseline, then evaluates how much this influences predictions. Of course, this means that it requires data to operate on - unlike static analysis methods that directly inspect the parameters of the model, feature ablation requires that the dataset be passed in as a parameter. Thus, we reserved a set of test data, separate from the training dataset, that we could use for model evaluation and interpretation.

\subsection{Adding The Transformer}
Transformers, as introduced in \cite{NIPS2017_3f5ee243}, have recently brought about significant advancements in time series analysis. Although groundbreaking in the field of large language models, where transformers are arranged in an encoder/decoder architecture to generate coherent text, the fundamental principle of self attention underlying transformers is applicable to a wide variety of time series tasks.

RNNs like LSTMs store context over time in a ``hidden state'' - after consuming each element $x$ of the input sequence, this hidden state is updated by passing it through a function $f(x, h) \rightarrow (y, h) $, a function that accepts an input $x$ and the current hidden state $h$, then returns an output $y$ and a new hidden state. The output of the RNN is the last $y$ produced - which means that the information that is used to compute the final output of the RNN is limited to the second-to-last hidden state and the last element. Although in theory, RNNs can operate on infinitely long sequences, the actual amount of context that can be used to compute the output of a given sequence element is limited to the amount of information that can be encoded by the hidden state vector. This results in a \textbf{temporal dependency}: the context to compute the output of any sequence element must come from sequence elements close to it. 

More damaging is the inherently sequential nature of RNNs. The computation of hidden state $h_k$ depends on $h_{k - 1}$. This unfortunate fact means that the computation of output $y_n$ requires the computation of the $n - 1$ hidden states before it, rendering parallelization impossible. We'd like a model that is capable of leveraging modern GPUs.

Transformers address both of these issues through a mechanism known as ``self attention''. Rather than sequentially computing hidden states / sequence outputs, self attention applies 3 separate transformations to the input, turning it into ``query'', ``key'', and ``value'' vectors (matrices, in practice). The query and key vectors are multiplied and transformed in a pairwise fashion, allowing the computation of a matrix called the ``attention weights''. Essentially, this matrix encodes how important each element of the input is to each element of the output. It's somewhat analogous to computing all of the hidden states in a single matrix multiplication, all at once. It can then be used to transform the ``value'' vector into a final output sequence. 

Since the attention weights are computed by running some transformation $f(a, b)$ on every single pair $(a, b)$, the context usable by each sequence element includes the entire sequence, with equal potential importance given to each element (the actual importance is the attention weight). This successfully breaks temporal dependency. Furthermore, since the attention weights are computed all at once in a single matrix multiplication, this breaks the sequential nature of RNNs and lends itself to massive parallelization. 

Therefore, we developed a transformer-based architecture to replace the LSTM. Generative applications of the transformer usually leverage both an encoder and a decoder, wherein previous sequences are used as input to the decoder in order to generate new sequences. However, since each pose sequence in our application is independent, we actually do not need a decoder at all and our model can perform with just an encoder.

\begin{table*}[hbt!]
     \centering
     \caption{DIP-IMU Model Performance On TotalCapture}
     \label{tab:dipimures}
     \begin{tabular}{lcccc}
     \hline
     \textbf{Method} & \(\mu_{ang}\) [deg] & \(\sigma_{ang}\) [deg] & \(\mu_{pos}\) [cm] & \(\sigma_{pos}\) [cm] \\
     \hline
     SOP             & 22.18               & 17.34                  & 8.39               & 7.57                 \\
     SIP             & 16.98               & 13.26                  & 5.97               & 5.50                 \\
     RNN (Dropout)   & 16.83               & 13.41                  & 6.27               & 6.32                 \\
     RNN (Acc)       & 16.07               & 13.16                  & 6.06               & 6.01                 \\
     RNN (Acc+Dropout) & 16.08             & 13.46                  & 6.21               & 6.27                 \\
     BiRNN (Dropout) & 15.86               & 13.12                  & 6.09               & 6.01                 \\
     BiRNN (Acc)     & 16.31               & 12.28                  & 5.78               & 5.62                 \\
     BiRNN (Acc+Dropout) & 15.85           & 12.87                  & 5.98               & 6.03                 \\
     BiRNN (after fine-tuning) & 16.84     & 13.22                  & 6.51               & 6.17                 \\
     \hline
     \end{tabular}
\end{table*}

Below, we have provided our model architecture as a list. Here, $N$ is the number of sensors the model uses - it will either be 24 or 6.
\begin{enumerate}
    \item Linear Layer: Transform from shape $N * (3*3 + 3)$ to $512$.
    \item Positional Encoding: Since transformers have no notion of position, augment each input with information about where it occurs in the sequence as per \cite{NIPS2017_3f5ee243}. Transforms from $512$ to $512$.
    \item Transformer Encoder: Stack of 6 identical layers, each accepting and returning $512$. Each layer implements multi-headed self attention with 4 heads, then runs the concatenated result through a feedforward network to transform it back to $512$.
    \item Linear Layer: Transform from $512$ back to $24 * (3*3)$.
\end{enumerate}

We trained this for 5 epochs on an RTX3090, where each epoch took around 4 minutes. The transformer architecture yields performance around 5 times as fast as the LSTM.

\section{Results and Discussion}

In this study, our objective was to attain higher pose estimation accuracy through a combination of data-driven IMU placement and a novel application of the transformer network architecture.

\subsection{Discussion of Model Performance}

As a result of our data-driven IMU placement strategy, both our LSTM and transformer-powered models achieve better performance on the TotalCapture \cite{TotalCapture_Trumble:BMVC:2017} dataset than DIP-IMU, the previous work in this domain \cite{huang_deep_2018}. See table \ref{tab:dipimures} for DIP-IMU's results on TotalCapture - the $\mu_{ang}$ captures local rotation errors for each type of model. The important value is $15.85$, which is the best local rotation error that DIP-IMU achieved on TotalCapture.

\begin{table}[ht]
     \centering
     \begin{tabular}{|l|l|}
     \hline
     Attribute       & Value    \\
     \hline
     crit\_type      & MSELoss() \\
     crit\_score     & 0.010    \\
     pos\_err        & 0.094    \\
     loc\_rot\_err   & 8.695   \\
     global\_rot\_err & 15.221  \\
     \hline
     \end{tabular}
     \caption{Transformer TotalCapture Performance With 24 Sensors}
     \label{tab:fulltransformertc}
\end{table}

\begin{table}[ht!]
     \centering
     \begin{tabular}{|l|l|}
     \hline
     Attribute       & Value    \\
     \hline
     crit\_type      & MSELoss() \\
     crit\_score     & 0.025    \\
     pos\_err        & 0.184    \\
     loc\_rot\_err   & 13.892   \\
     global\_rot\_err & 40.588  \\
     \hline
     \end{tabular}
     \caption{BiRNN TotalCapture Performance With 24 Sensors}
     \label{tab:fullbirnntc}
\end{table}

To evaluate baseline performance - i.e. the best possible rotational error that our models could achieve - we ran evaluations on the models trained on 24 IMUs. Table \ref{tab:fullbirnntc} captures the evaluations for the LSTM, while table \ref{tab:fulltransformertc} captures the evaluations for the transformer. Therefore, the ideal rotational error we hope to achieve would be 13.892 degrees with an LSTM, and 8.695 degrees with a transformer. 

That said, tables \ref{tab:optimbirnntc} and \ref{tab:optimtransformertc} reveal the performance of the optimized LSTM and transformer models, respectively. Notice that the optimized LSTM has a local rotation error of 13.018 on TotalCapture, while the optimized transformer has a local rotation error of 12.916. Both models yield significantly better performance than TotalCapture's solution, while the transformer, as expected, performs slightly better than the LSTM.

\begin{table}[hbt!]
     \centering
     \caption{BiRNN TotalCapture Performance With 6 Sensors}
     \label{tab:optimbirnntc}
     \begin{tabular}{|l|l|}
     \hline
     Attribute       & Value    \\
     \hline
     crit\_type      & MSELoss() \\
     crit\_score     & 0.023    \\
     pos\_err        & 0.155    \\
     loc\_rot\_err   & 13.018   \\
     global\_rot\_err & 29.308  \\
     \hline
     \end{tabular}
\end{table}

\begin{table}[hbt!]
     \centering
     \caption{Transformer TotalCapture Performance With 6 Sensors}
     \label{tab:optimtransformertc}
     \begin{tabular}{|l|l|}
     \hline
     Attribute       & Value    \\
     \hline
     crit\_type      & MSELoss() \\
     crit\_score     & 0.021    \\
     pos\_err        & 0.125    \\
     loc\_rot\_err   & 12.916   \\
     global\_rot\_err & 25.088  \\
     \hline
     \end{tabular}
\end{table}

Do remember, however, that even though the accuracy delta between the LSTM and transformer isn't that large, the transformer's massively parallel nature meant that it only took a fifth of the time to train.

\subsection{Discussion of ``Optimal'' IMU Locations}

Just looking at model evaluations hides some really interesting parts of the story. When we ran feature ablation, we observed that the most relevant IMUs were quite dataset dependent, in addition to the fact that the transformer and LSTM seemed to prefer vastly different sets of IMUs. 

\begin{figure}[hbt!]
     \centering
     \includegraphics[width=1\linewidth]{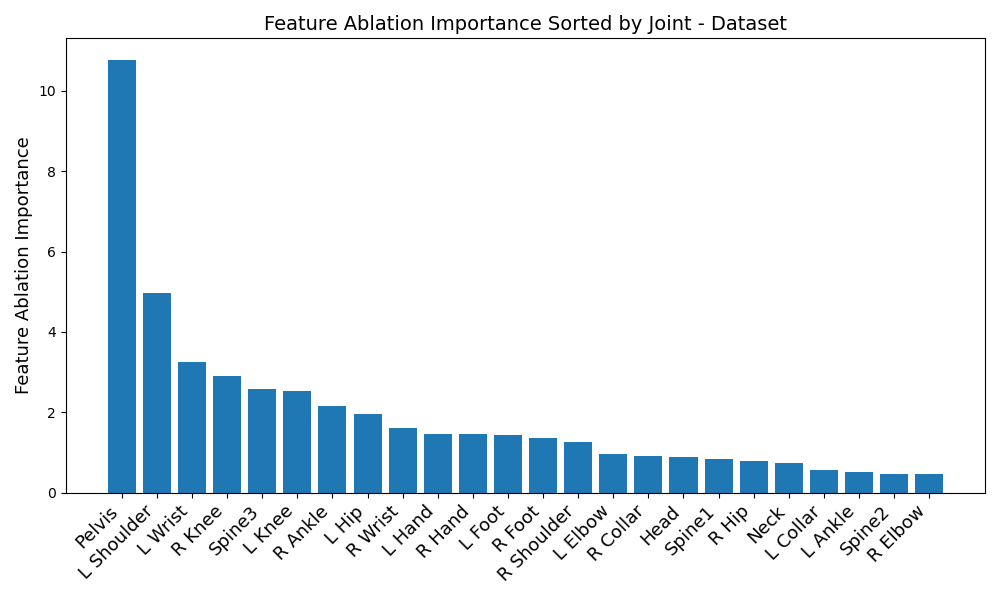}
     \caption{Global BiRNN Feature Ablation}
     \label{fig:globalbirnnablation}
\end{figure}

\begin{figure}[hbt!]
     \centering
     \includegraphics[width=1\linewidth]{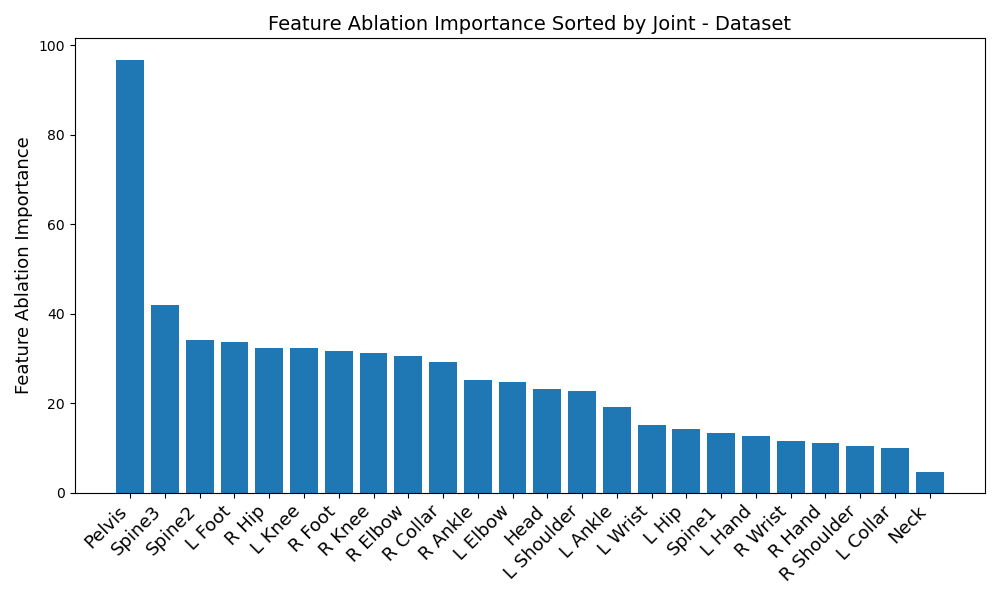}
     \caption{Global Transformer Feature Ablation}
     \label{fig:globaltransformerablation}
\end{figure}

Start off by looking at figures \ref{fig:globalbirnnablation} and \ref{fig:globaltransformerablation}. Immediately, a disparity in preferred IMUs emerges - the BiRNN prefers the pelvis, left shoulder, left wrist, right knee, upper spine, and left knee, in that order. However, the transformer prefers the pelvis, upper spine, middle spine, left foot, right hip, and left knee, in that order. Although there are certainly some overlaps - namely the pelvis, upper spine, and left knee, three of the joints differ. 

\begin{figure}[hbt!]
     \centering
     \includegraphics[width=1\linewidth]{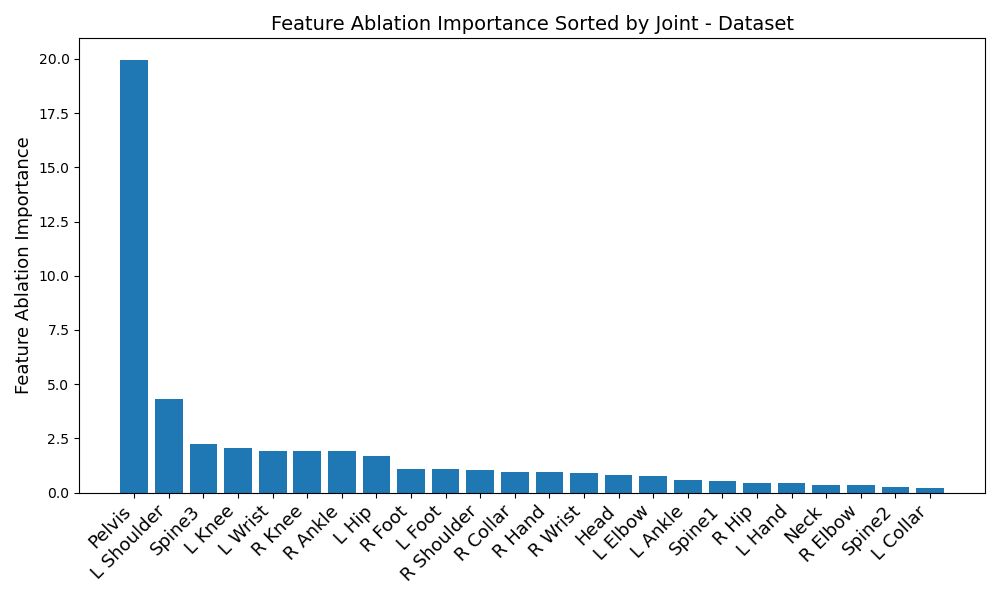}
     \caption{ACCAD BiRNN Feature Ablation}
     \label{fig:accbirnnablation}
\end{figure}

\begin{figure}[hbt!]
     \centering
     \includegraphics[width=1\linewidth]{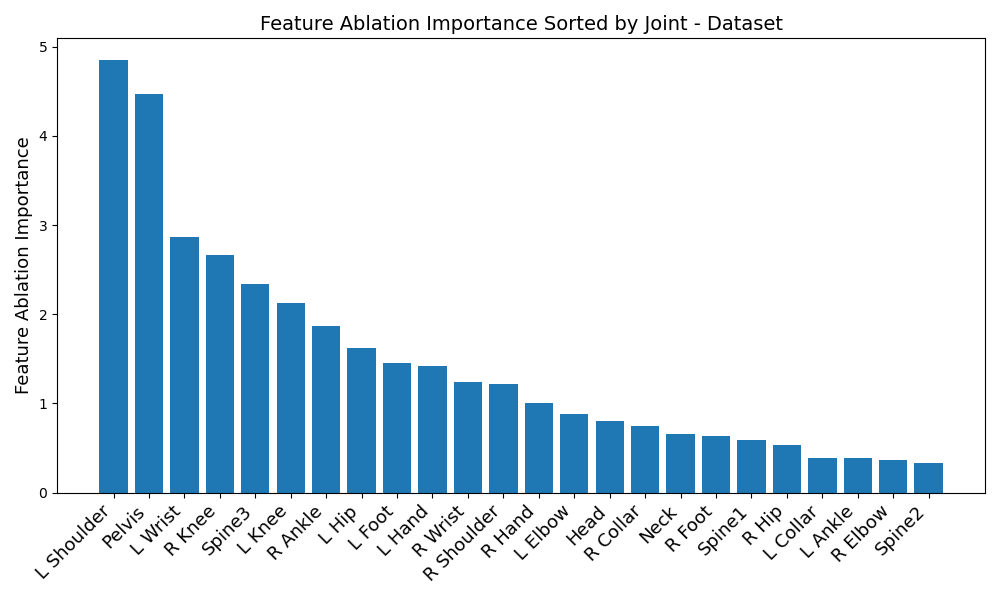}
     \caption{BioMotionLab\_NTroj BiRNN Feature Ablation}
     \label{fig:biobirnnablation}
\end{figure}

\begin{figure}[hbt!]
     \centering
     \includegraphics[width=1\linewidth]{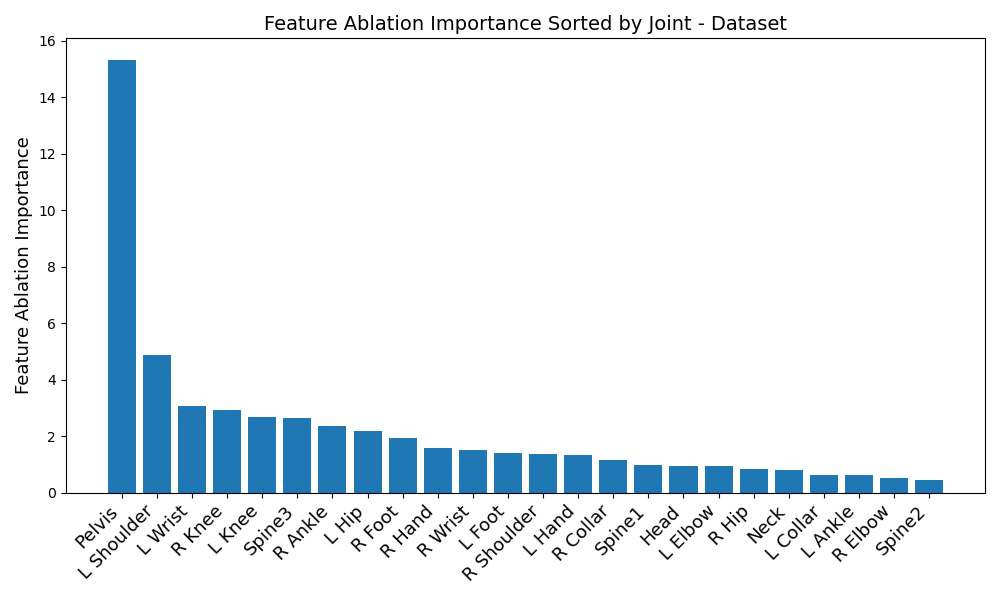}
     \caption{CMU BiRNN Feature Ablation}
     \label{fig:cmubirnnablation}
\end{figure}

\begin{figure}[hbt!]
     \centering
     \includegraphics[width=1\linewidth]{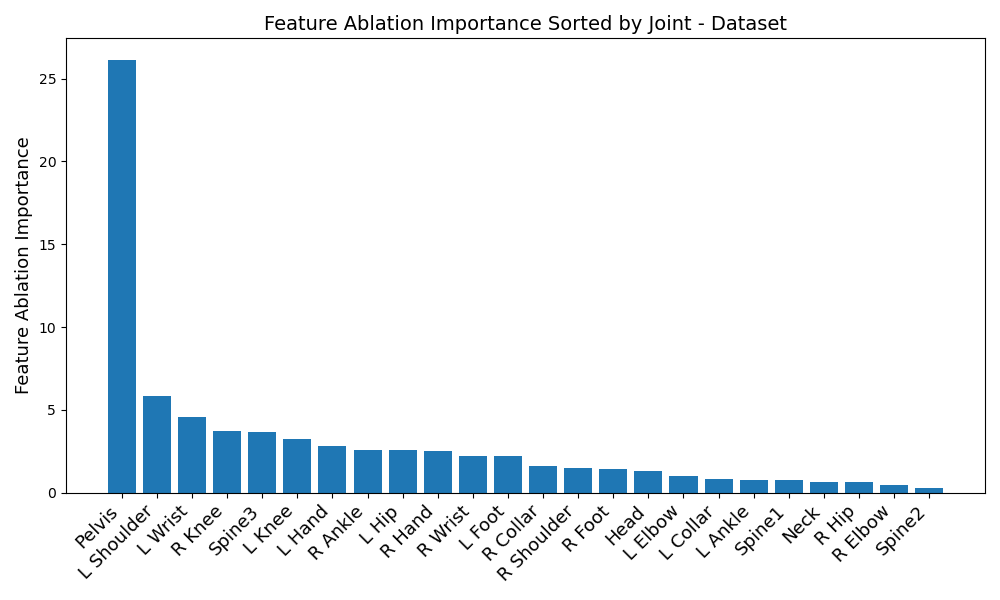}
     \caption{TotalCapture BiRNN Feature Ablation}
     \label{fig:tcbirnnablation}
\end{figure}

Another disparity appears on a per-dataset basis. Figures \ref{fig:tcbirnnablation}, \ref{fig:accbirnnablation}, \ref{fig:biobirnnablation}, and \ref{fig:cmubirnnablation} reveal the feature ablation results on the TotalCapture \cite{TotalCapture_Trumble:BMVC:2017}, ACCAD \cite{ACCAD}, CMU \cite{cmuWEB}, and BioMotionLab\_Ntroj \cite{BMLrub} datasets. Although each dataset does seem to yield the same set of optimal IMUs as the globally optimal BiRNN IMUs, they occur with different degrees of importance.
\begin{enumerate}
    \item ACCAD: Pelvis, left shoulder, upper spine, left knee, left wrist, right knee
    \item BioMotionLab\_Ntroj: Left shoulder, pelvis, left wrist, right knee, upper spine, left knee.
    \item CMU: Pelvis, left shoulder, left wrist, right knee, left knee, upper spine
    \item TotalCapture: Pelvis, left shoulder, left wrist, right knee, upper spine, left knee
\end{enumerate}

A similar sort of disparity occurs with the transformer - take a look at the listing below:
\begin{enumerate}
    \item ACCAD: Pelvis, upper spine, left knee, left foot, right knee, right elbow
    \item BioMotionLab\_NTroj: Upper spine, left foot, middle spine, left knee, right foot, right hip
    \item CMU: Pelvis, upper spine, left foot, right knee, middle spine, right hip
    \item TotalCapture: Pelvis, upper spine, middle spine, left foot, left knee, right foot
\end{enumerate}

\begin{figure}[hbt!]
     \centering
     \includegraphics[width=1\linewidth]{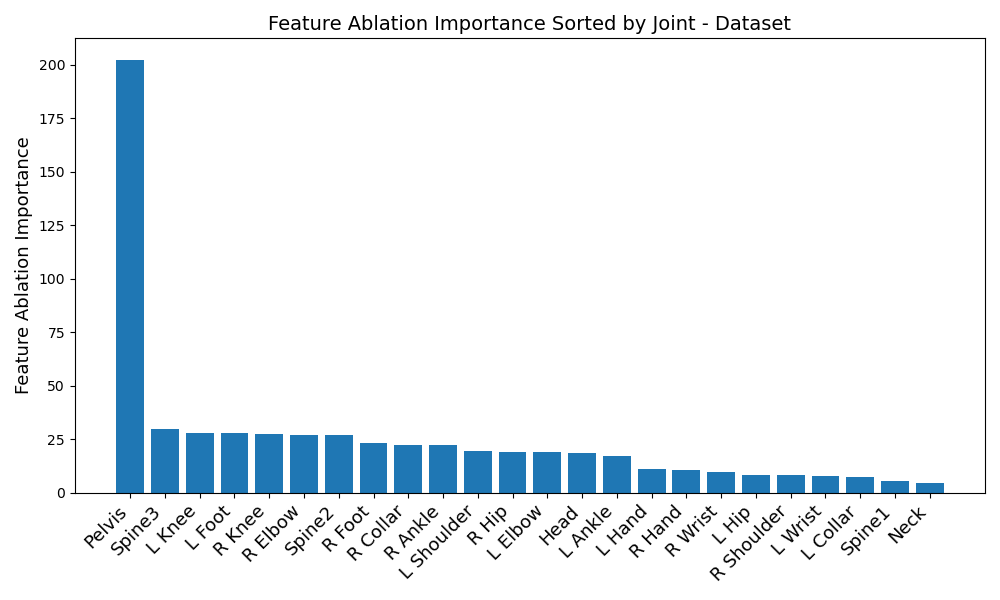}
     \caption{ACCAD Transformer Feature Ablation}
     \label{fig:acctransformerablation}
\end{figure} 

\begin{figure}[hbt!]
     \centering
     \includegraphics[width=1\linewidth]{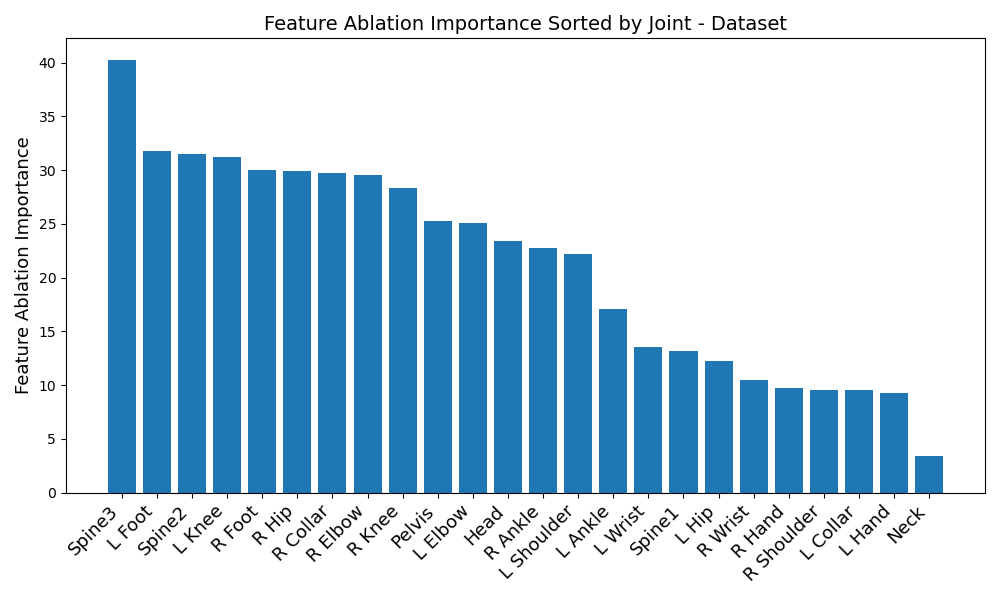}
     \caption{BioMotionLab\_NTroj Transformer Feature Ablation}
     \label{fig:biotransformerablation}
\end{figure} 

\begin{figure}[hbt!]
     \centering
     \includegraphics[width=1\linewidth]{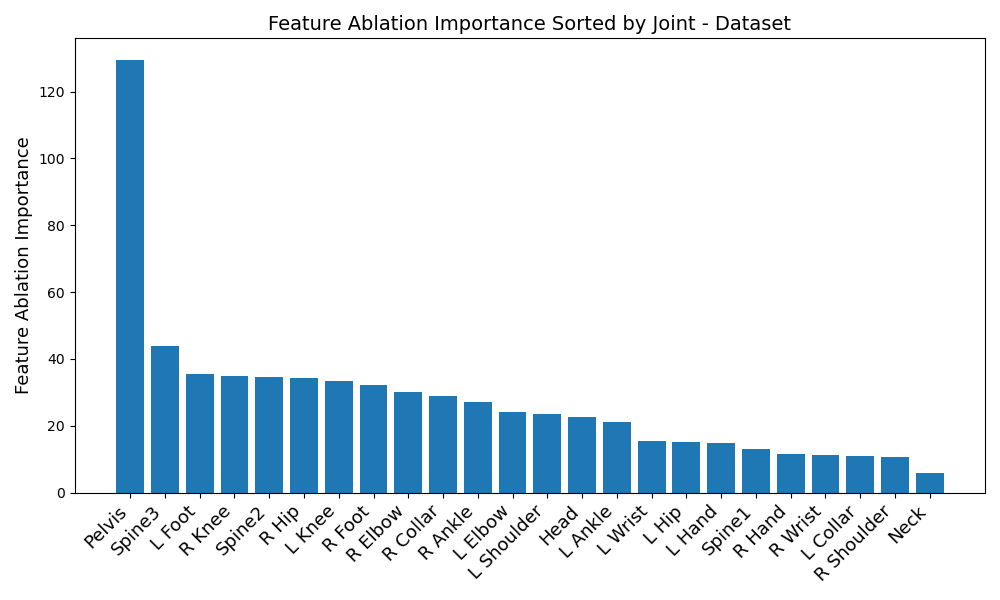}
     \caption{CMU Transformer Feature Ablation}
     \label{fig:cmutransformerablation}
\end{figure} 

\begin{figure}[hbt!]
     \centering
     \includegraphics[width=1\linewidth]{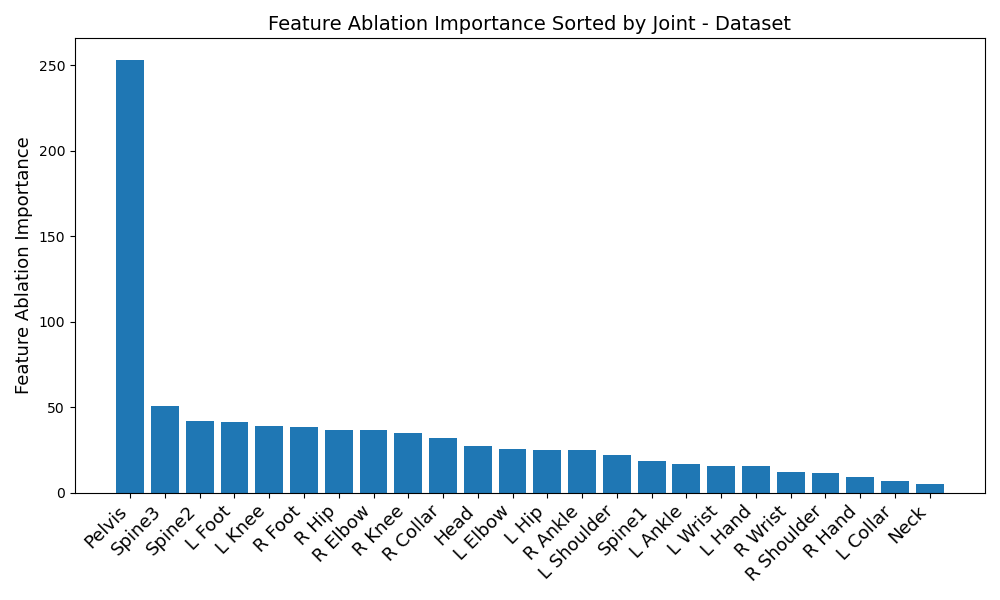}
     \caption{TotalCapture Transformer Feature Ablation}
     \label{fig:tctransformerablation}
\end{figure}

This time, many individual joints differ as well, rather than just being reordered. This could point to the transformer's potential for extracting finer nuances of the data - but in either case, reveals that finding globally optimal IMUs in all situations might not be possible. The vastly differing results between datasets and model architectures reveals that IMUs should be carefully chosen in a data-driven manner with regards to individual activity types. Although this paper does try to approximate an optimal set of IMUs for each model architecture, the true result of our study is that the ``optimal'' set of IMUs for pose estimation should be tailored to the situation.

\section{Conclusion}
\new
Our exploration into the use of sparse IMU configurations for human pose estimation has culminated in a robust methodology that outpaces the capabilities of conventional biRNN models. By employing a data-driven strategy to determine the optimal placement of IMUs on the body and leveraging the computational strengths of the transformer architecture, we have achieved notable advancements in pose reconstruction accuracy from IMU data.

Our findings underscore the significance of IMU placement, revealing that the strategic selection of sensor locations is crucial and highly contingent on the dataset and activity type. The contrasting preferences between LSTM and transformer models for certain IMUs, along with the variations across different datasets, illuminate the complexity and situational dependency of optimal sensor placement. This work emphasizes that there is no one-size-fits-all solution; instead, IMU placement must be tailored to the specific context and objectives of the application.

Furthermore, the rapid training times and parallel processing capabilities of the transformer model present compelling advantages over traditional LSTM models. This efficiency, combined with the transformer's proficiency in time-series analysis, positions our approach as a scalable and practical solution for real-world applications.

In summary, this study not only contributes a novel methodological framework for IMU-based pose estimation but also provides key insights into the nuanced relationships between IMU data and human pose. Our results pave the way for future research in this domain, encouraging the development of more adaptive, efficient, and precise pose estimation systems that capitalize on the ubiquity and versatility of IMUs in consumer devices.

% \newpage
\bibliographystyle{acm}
% \nocite{*}
\bibliography{imuoptimize}

%\balance

\end{sloppypar}

\newpage

\end{document}